\documentclass[10pt,technote,compsoc]{IEEEtran}
\ifCLASSOPTIONcompsoc
  \usepackage[nocompress]{cite}
\else
  \usepackage{cite}
\fi
\ifCLASSINFOpdf
\else
\fi
\usepackage{url}
\usepackage{amsmath,amssymb,amsthm}
\usepackage{multirow}
\usepackage{booktabs}
\usepackage{graphicx,subfigure}
\usepackage[dvipsnames,table]{xcolor}
\graphicspath{ {./fig/} }

\usepackage{tikz}
\usetikzlibrary{decorations.pathmorphing} 
\usetikzlibrary{fit}					
\usetikzlibrary{backgrounds}	

\newcommand{\na}[1]{#1}

\begin{document}
%
\title{Sum-Product Networks\\ for Sequence Labeling}
%
%
%
%

\author{Martin~Ratajczak,
        Sebastian~Tschiatschek,
        and~Franz~Pernkopf,~\IEEEmembership{Senior~Member,~IEEE}
\IEEEcompsocitemizethanks{\IEEEcompsocthanksitem Martin~Ratajczak and Franz~Pernkopf are associated with the Department
of Signal Processing and Speech Communication Laboratory, Graz University of Technology, 8010 Graz, Inffeldgasse 16c.\protect\\
E-mail: see http://www.spsc.tugraz.at/people
}
\IEEEcompsocitemizethanks{\IEEEcompsocthanksitem Sebastian~Tschiatschek is a researcher at Microsoft Research, Cambridge, United Kingdom.\protect\\
E-mail: see https://www.microsoft.com/en-us/research/people
}
\thanks{This work was supported by the Austrian Science Fund (FWF) under
the project number P25244-N15 and P27803-N15. Furthermore, we
acknowledge NVIDIA for providing GPU computing resources.}}

\IEEEtitleabstractindextext{%
\begin{abstract}
We consider higher-order linear-chain conditional random fields (HO-LC-CRFs) for sequence modelling, and use sum-product networks (SPNs) for representing higher-order input- \emph{and} output-dependent factors. SPNs are a recently introduced class of deep models for which exact \emph{and} efficient inference can be performed. By combining HO-LC-CRFs with SPNs, expressive models over both the output labels and the hidden variables are instantiated while still enabling efficient exact inference. Furthermore, the use of higher-order factors allows us to capture relations of multiple input segments and multiple output labels as often present in real-world data. These relations can \emph{not} be modeled by the commonly used first-order models and higher-order models with local factors including only a single output label. We demonstrate the effectiveness of our proposed models for sequence labeling. In extensive experiments, we outperform other state-of-the-art methods in optical character recognition and achieve competitive results in phone 
classification.
\end{abstract}

}

\maketitle

\IEEEdisplaynontitleabstractindextext

%
\IEEEpeerreviewmaketitle


\section{Introduction}
In \emph{sequence labeling}, a given input sequence $\bold{x}$, e.g.\ a time series, is mapped to an output label sequence $\bold{y}$.
\emph{Maximum entropy Markov models (MEMMs)}~\cite{McCallum2000} and \emph{Linear-chain conditional random fields (LC-CRFs)}~\cite{Lafferty2001} are established discriminative probabilistic models for sequence labeling. For instance, they have been successfully used for speech recognition~\cite{Gunawardana2005}, optical character recognition and natural language processing~\cite{Ye2009}.
Due to several advantages, LC-CRFs achieve better performance compared to their generative counterparts, i.e.\ hidden Markov models (HMMs)~\cite{Gunawardana2005}. While LC-CRFs are normalized over the whole sequence, thereby counteracting the \emph{label bias problem}, MEMMs are normalized locally. Nevertheless, MEMMs are of interest in various applications as they can be easily extended to arbitrary long histories and have lower time complexity in training.


First-order LC-CRFs typically consist of transition factors, modeling the relationship between two consecutive output labels, and local factors, modeling the relationship between input observations (usually a sliding window over input frames) and one output label. But LC-CRFs are not limited to these types of factors: 
\emph{Higher-order LC-CRFs (HO-LC-CRFs)} allow for arbitrary input-independent (such factors depend on the output labels only)~\cite{Ye2009} and input-dependent (such factors depend on both the input and output variables) higher-order factors~\cite{Qian2009,Lavergne2011}. That means both types of factors can include more than two output labels.\footnote{Clearly, the Markov order of the largest factor (on the output side) dictates the order of the LC-CRF.} Higher-order input-dependent factors can model relations of the input and multiple output labels, which are often present in real-world data.

It is common practice to represent the higher-order factors by linear functions which can reduce the model's expressiveness~\cite{Ratajczak2015a}.
In the case of first-order input-dependent factors, a widely used approach to overcome this limitation, is to represent non-linear dependencies by parametrized models and to learn these models from data. Several approaches have been suggested to parametrize \emph{first-order} factors in LC-CRFs, mainly kernel methods~\cite{Lafferty2004} and \emph{neural models}~\cite{Larochelle2008,Peng2009,Prabhavalkar2010,Maaten2011,Ratajczak2014}.
In summary, most work in the past focused either on (a) higher-order factors represented by simple linear models, or (b) first-order input-dependent non-linear factors mapping an input sub-sequence to one output label. A noteworthy exception is the \emph{neural higher-order LC-CRF (NHO-LC-CRF)}~\cite{Ratajczak2015a} which uses multi-layer perceptron networks (MLPs) to model input-dependent higher-order factors. 

Indeed, higher-order CRFs increase the model complexity as the number of features grows exponentially with the number of the output variables considered in higher order factors~\cite{Stewart2008}. Consequently, to avoid overfitting, the amount of training data has to be sufficiently large for training. Alternatively,  a suitable representation may reduce the overfitting problem as it has been observed for neural networks~\cite{Ratajczak2015a}.\\ 
%
In this work, we explore a specific type of sum-product networks (SPNs)~\cite{Poon2011,Gens2012,Peharz2017}  \emph{and} use it for modelling higher-order input-dependent factors in LC-CRFs. SPNs~\cite{Poon2011} enable one to perform efficient \emph{and} exact training of deep models with many layers of hidden variables. They attracted attention as the \emph{discriminative SPN} outperformed deep neural networks and other methods on a difficult image classification task~\cite{Gens2012}. However, their performance on other discriminative tasks is largely unexplored.
In contrast to typical deep RBMs~\cite{Hinton2002}, the considered SPNs go beyond pairwise factors 
and model the relationship between variables in multiple 
layers from the top to the lowest layer.
Note that in general, exact inference in such models is intractable.
\\

\noindent Our main contributions are:\\
(i) We explore an \emph{extension of LC-CRFs and MEMMs} by deep input-dependent factors represented by SPNs. The LC-CRF with SPN factors represents a probabilistic model over visible and hidden variables. This model allows for efficient and exact inference (e.g.\ computation of the marginals of the hidden variables and the output labels).\\
(ii) We propose the use of SPNs as higher-order factors in LC-CRFs, enabling to model rich dependencies between several observations and several output labels. \\
(iii) We demonstrate the effectiveness of our models in extensive \emph{sequence labeling} experiments, achieving competitive performance for phone classification and handwriting recognition. 


The remainder of this paper is structured as follows: 
In Section~\ref{sec:related_work} we review related work.
In Section~\ref{sec:models} we introduce a specific type of SPNs for classification and discuss their usage within LC-CRFs as well as MEMMs for sequence labeling. 
In Section~\ref{sec:results} we evaluate these models on two challenging sequence labeling tasks, i.e.\ handwriting recognition and phone classification.
Finally, Section~\ref{sec:conclusion} concludes the paper.


\section{Other Related Work}
\label{sec:related_work}

Discriminative SPNs have been introduced in~\cite{Gens2012}.
Our work differs in several points. First, we formulate our model in a different way
which is not based on Darwiche's network polynomial \cite{Darwiche2000,Darwiche2003}.
Second, we utilize message passing to compute the model's marginal probabilities in contrast to back-propagation
\cite{Poon2011,Gens2012}.
Last but not least, we aimed at sequence labeling in contrast to single label classification task.

In previous work, deep architectures have been used in LC-CRFs.
Some of them use generative and unsupervised pre-training  on the input data to improve the generalization.
One of them is using deep belief networks (DBNs),~i.e. RBMs are trained layer by layer.
The resulting DBNs are then transformed into a multi-layer neural network
and plugged into the LC-CRF \cite{Do2010}.
Finally, the whole model, i.e.\ the LC-CRF and the deep model, is fine-tuned by back-propagation. Other approaches, such as conditional neural fields (CNFs) \cite{Peng2009} and multi-layer CRFs \cite{Prabhavalkar2010}, propose to jointly optimize multi-layer neural networks and LC-CRFs directly using the conditional likelihood criterion based on error back-propagation.

There is only a small number of models as our proposed model which represent a probability distribution over the output
\emph{and} the hidden variables and allow for exact and efficient inference.
A well known example is the Gaussian mixture model (GMM) which has been applied extensively for many years in conjunction with HMMs and LC-CRFs~\cite{Fosler2013} because of its scalability. However, it is known that the performance of GMMs is inferior to neural networks.
Another approach is the \emph{hidden-unit conditional random field} (HU-CRF)~\cite{Maaten2011} which extends the LC-CRF by replacing the local factors with the \emph{discriminative RBM} (DRBM) \cite{Larochelle2008}. Unfortunately, the HU-CRF~\cite{Maaten2011} is limited to a single hidden layer and the local factors simply map to a single output label. Although it can be interpreted as a neural network, it is a probabilistic undirected graphical model (UGM) supporting exact and efficient inference.

Furthermore, we emphasize the relation of the considered SPNs to \emph{context-specific undirected graphical models} with \emph{higher order factors}~\cite{Tarlow2010,Nyman2013}.

%


\section{Sum-Product Networks}
\label{sec:models}

\begin{figure}[th!]
\centering
\subfigure[]{\includegraphics[width=0.52\columnwidth]{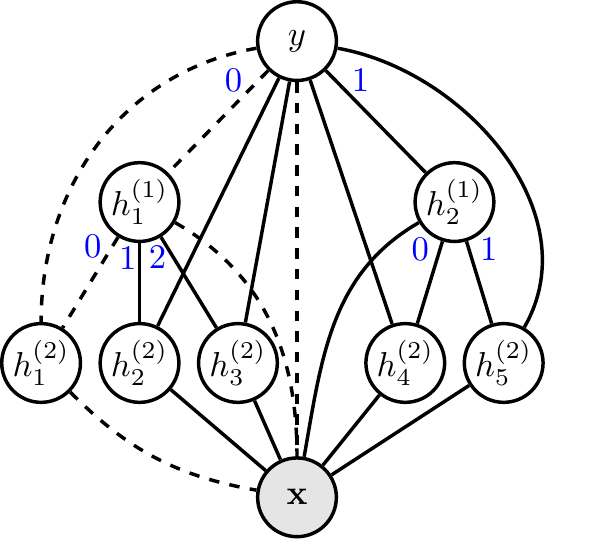}}
\subfigure[]{\includegraphics[width=0.36\columnwidth]{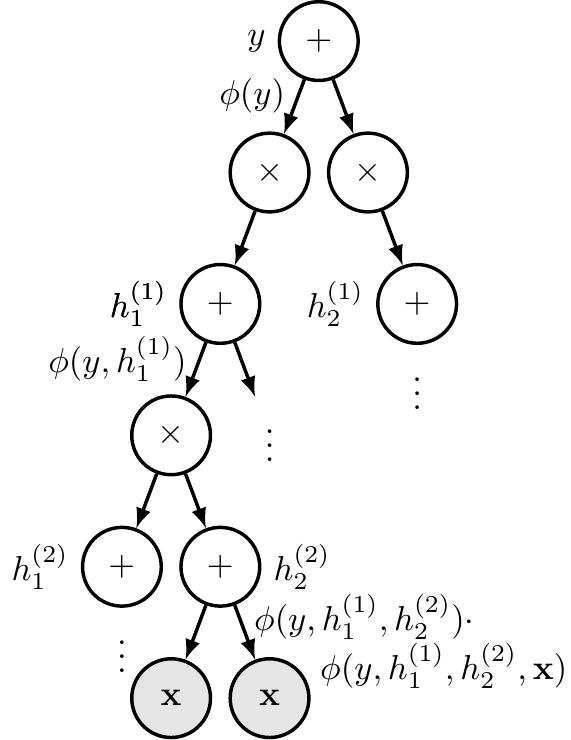}}
\caption{SPN classifier represented as (a) conditional undirected graphical model and as (b) sum-product network. Dashed edges indicate a maximal clique.}
\label{fig:dhcrf}
\end{figure}

We introduce an SPN classifier represented as a conditional undirected graphical model and show the equivalence between the UGM and SPN representations. In Section \ref{sec:memm} and \ref{sec:lc-crf_cd-dhcrf} we integrate this model into MEMMs and LC-CRFs and name them SPN-MEMM and SPN-CRFs, respectively.

\subsection{Sum-Product Network Classifier}
\label{sec:cd-dhcrf}

\subsubsection{Model Definition} The probability distribution of an UGM is defined as
the product over a set of maximal clique factors $\phi_k(\cdot)$ and the corresponding normalization constant. In this way, we define our model as
\begin{align}
  p(y,\textbf{h}|\textbf{x}) = \frac{\prod_k \phi_k(y,\textbf{h},\textbf{x})}{Z(\textbf{x})} \text{\,} \label{eq:prob}
\end{align} 
over the output variable $y$ (class label) \emph{and} a set of hidden variables $\textbf{h}$ given a set of input variables $\textbf{x}$. The set of hidden variables $\textbf{h}= \cup_{l=1}^L \textbf{h}^{(l)}$ is the union of the hidden variables $\textbf{h}^{(l)}$ over $L$ hidden layers and $Z(\textbf{x})$ is the
partition function. 
The posterior probability distribution of the output variable $y$ can be computed by marginalizing over the hidden variables $\textbf{h}$, i.e.
\begin{align}
p(y|\textbf{x})=\frac{Q(y,\textbf{x})}{Z(\textbf{x})} \text{\,,}
\end{align}
where $Q(y,\textbf{x})=\sum_{\textbf{h}}\prod_k \phi_k(y,\textbf{h},\textbf{x})$ and $Z(\textbf{x}) = \sum_y Q(y,\textbf{x})$.
Without further assumptions, computing the partition function is intractable. Therefore, we restrict our model to a specific model structure to enable efficient inference.

\subsubsection{Model Structure} 
\label{sec:model-structure}
An instance of our model with two hidden layers is shown in Figure \ref{fig:dhcrf}a, represented as an undirected graphical model.
The nodes in the graph represent input variables $\textbf{x}$, 
multiple layers of hidden variables $\textbf{h}$ and one output variable $y$. The edges represent direct dependencies between variables.
The restrictions in our model are: First, no edges connect the hidden variables within the same layer. 
Second, hidden variables must not have edges to more than one hidden variable in the layer immediately above (its immediate parent).
Third, hidden variables connect not only to their immediate parents but also to the parents of their immediate parents and so on. 
This way, our model represents \emph{higher order factors} going beyond pairwise factors usually used in RBMs.
For instance, in Figure \ref{fig:dhcrf}a, the  set of variables $\{y,h_1^{(1)},h_1^{(2)},\textbf{x}\}$ forms a maximal clique in the UGM, i.e.\ a fully connected subgraph. The factor for this clique is decomposed and modeled by a bias factor $\phi(y)$, a pairwise factor $\phi(y,h_1^{(1)})$ and \emph{higher order factors} $\phi(y,h_1^{(1)},h_1^{(2)})$ and  $\phi(y,h_1^{(1)},h_1^{(2)},\textbf{x})$.

\noindent According to our model structure, we can define paths of variables $\mathcal{S}$ rooted in $y$ and leading to $\textbf{x}$ via $h_i^{(1)} \in \textbf{h}^{(1)}, \dots, h_j^{(L)} \in \textbf{h}^{(L)}$, i.e.
\begin{equation}
 \mathcal{S} = ( S_0 = y, S_1 = h_i^{(1)} , \dots, S_{L} = h_j^{(L)}, S_{L+1}=\textbf{x} ).
\end{equation}
According to the model structure, every such path $\mathcal{S}$ forms a maximal clique $\phi(\mathcal{S})$.

\noindent As exemplified above, we assume that the maximal cliques $\phi(\mathcal{S})$ decompose into factors $\phi(S_0), \phi(S_{0:1}), \ldots, \phi(S_{0:L+1})$, where $S_{0:l}$ denotes the set of variables $\{ S_0, S_1, \ldots, S_l \}$. The product in~\eqref{eq:prob} is computed over the following set of factors  
\begin{equation}
 \{ \phi(S_{0:l}) \mid \forall \mathcal{S} \; \forall l=0,\ldots,L+1\}. 
\end{equation}

\subsubsection{Sum-product Form} The summation over the hidden variables in $Q(y,\textbf{x})$ can be reordered.
In particular, the assumed factorization of the maximal cliques allows the computation of $Q(y,\textbf{x})$ as 
\begin{align}
& Q(y,\textbf{x})=\phi(S_0)
\prod_{S_1 \in\mathbf{S}_1(S_0)} \sum_{\textnormal{val}(S_1)} \phi(S_{0:1}) \label{eq:Qsp} \\
&\quad \prod_{S_2 \in\mathbf{S}_2(S_{0:1})} \sum_{\textnormal{val}(S_2)} \phi(S_{0:2}) \; \cdots \nonumber\\
&\quad\prod_{S_{L} \in\mathbf{S}_{L}(S_{0:L-1})} \sum_{\textnormal{val}(S_{L})} \phi(S_{0:L})
\phi(S_{0:L+1}), \nonumber
\end{align}
where $\mathbf{S}_{p}(S_{0:p-1})$ denotes the set of hidden variables that immediately follow  in any path $\mathcal{S}$ whose first $p$ variables are $S_{0:p-1}$ and $\textnormal{val}(S_p)$ denotes the set of possible values of variable $S_p \in \mathbf{S}_{p}(S_{0:p-1})$.

In~\eqref{eq:Qsp}, products and weighted summations are alternated, which avoids an exhaustive summation over the whole state space.
The computation of the function $Q(y,\textbf{x})$ and the partition function $Z(\textbf{x})$
can be represented by a \emph{sum-product network} \cite{Poon2011} as illustrated in Figure \ref{fig:dhcrf}b. Weighted summations are represented as sum nodes, products as product nodes and the input variables as filled leave nodes.

\subsubsection{Model Parametrization} We parametrize our model as a log-linear model which is optimal with regard to the maximum entropy criterion under moment constraints~\cite{Berger1996}. Thus, the probability distribution of the model posterior is specified by the Gibbs distribution
\begin{align}
  p(y,\textbf{h}|\textbf{x}) = \frac{\exp(\sum_k w_k f_k(y,\textbf{h},\textbf{x}))}{Z(\textbf{x})}. \nonumber
\end{align}
The higher order factors $\phi_k(\cdot)=\exp(w_k\,f_k(\cdot))$ have corresponding weights $w_k$ and  feature functions $f_k(\cdot)$. To explain the precise form of the feature functions we considered, we introduce the following notation. By $S_{0:L}^o$ we represent an encoding of the states of the variables on the path $\mathcal{S}$ ending at layer $L$ in $S_L=o$. For instance, for the model in Figure~\ref{fig:dhcrf}, the path $\mathcal{S}=(y, h_1^{(1)}, h_1^{(2)}, \textbf{x})$ and assuming that variable $y$ can take values in $\{0,1\}$, $h_1^{(1)}$ can take values in $\{0,1,2\}$ etc., $S_{0:L}^o=[0, 0, o]$ (the hidden variables except for the ones in the last layer are encoded according to the blue state values in  Figure~\ref{fig:dhcrf}; $S_0=y=0,S_1=h_1^{(1)}=0,S_2=S_L=h_1^{(2)}=o$). 
Furthermore, let $S_{0:l}(y, \textbf{h})$ represent a vector containing the state values of the variables $S_{0:l}$ on the path $\mathcal{S}$ as given by the instantiatons $y$ and $\textbf{h}$.
Then, the feature functions involving the input variables $\textbf{x}$ are
$f_{S_{0:L}^o}(y,\textbf{h},\textbf{x})=\delta(S_{0:L}^o,S_{0:L}(y,\textbf{h})) \hat{f}_{S_{0:L}^o}(\textbf{x})$, where $\delta(\cdot,\cdot)$ is the indicator function and $\hat{f}_{S_{0:L}^o}(\textbf{x})$ is an arbitrary feature function---in our experiments we used log-linear feature functions $\textbf{w}_{S_{0:L}^o}^T\textbf{x}$. For the remaining layers $l \in \{0, \ldots, L\}$, the feature functions are $f_{S_{0:l}}(y, \textbf{h})=\delta(S_{0:l},S_{0:l}(y,\textbf{h}))$, where $S_{0:l}$ is defined analogeously to $S_{0:L}^o$ but restricted to the part $0:l$ of the path.

\subsubsection{Model Optimization} The model weights $\textbf{w}=(w_k)$ are optimized to maximize the logarithm of the conditional likelihood 
over the training set, i.e.
\begin{equation}
F(\textbf{w},\mathcal{D}) = \sum_{n=1}^N \log p(y_n|\textbf{x}_n), \nonumber
\end{equation}
where $\mathcal{D} = \{(y_1, \textbf{x}_1), \ldots, (y_N,\textbf{x}_N)\}$ is a given labeled training set drawn i.i.d.\ from an unknown data distribution.
To optimize the objective by first-order gradient ascent methods, we need to compute the partial derivatives of $F(\textbf{w},\mathcal{D})$ with respect to the weights. 
These gradients can either be computed using tools for automatic differentiation as nowadays commonly provided by deep learning frameworks~\cite{bergstra2010theano}, or by using the the results of~\cite{Gens2012}.

\subsubsection{Time Complexity} The time complexity for marginalization and gradient computation is $\mathcal{O}(Y I_f (IH)^L )$ assuming in each layer $l$ equal cardinality $H$ of the state space of the hidden variables 
and $I$ hidden variables per parent, where $Y$ is the number of class labels and $I_f$ is the number of feature functions in the layer $L+1$.
The time complexity is exponential in the number of hidden layers $L$ but is polynomial in $I$ and $H$.


\subsection{Sum-Product Networks for Maximum Entropy Models}
\label{sec:memm}
In this section, we extend higher order MEMMs by SPN local factors (SPN-MEMM). 
%
Higher-order MEMMs of order $M$ model the conditional probability of one label $y_t$ at sequence index $t$ given the $M-1$ previous labels $g_t = y_{t-M+1:t-1}$ and the observed sequence $\textbf{x}_{1:T}$ by
\begin{align}
  p(y_t |g_t,  \textbf{x}_{1:T}) = \frac{ \phi(y_t, \textbf{x}_{1:T}) \phi(g_t, y_t) }{Z(g_t, \textbf{x}_{1:T})},
\label{eq:pmemm}
\end{align}
where $T$ is the sequence length. The relationship between the label history $g_t$ and $y_t$ is modeled by transition factors $\phi(g_t,y_t)$. Further, the relationship between the input variables and labels
at sequence index $t$ is described by local factors $\phi(y_t, \textbf{x}_{1:T})$ modeled as SPNs.
MEMMs are locally normalized, i.e.\ 
$Z(g_t, \textbf{x}_{1:T}) = \sum_{y_t} \phi(y_t, \textbf{x}_{1:T}) \phi(g_t, y_t).$
The conditional probability of the sequence labels $y_{1:T}$ given the observed sequence $\textbf{x}_{1:T}$ is
\begin{align}
  p(y_{1:T} | \textbf{x}_{1:T}) = \prod_{t=1}^T p(y_t |g_t,  \textbf{x}_{1:T}).
\end{align}
We use distant bigram features $f_k(g_{t,m})=\delta(g_{t,m},y_{t-m})$ for all $m=1,\ldots,M-1$ previous labels where $k=y_{t-m}$ to model the transition factors
\begin{align}
  \phi(g_t,y_t)     &= \exp\left( \sum_{m=1}^{M-1} w_{g_{t,m}, y_t} \right).
\end{align}

\noindent The most probable sequence $
\hat{y}_{1:T} = \underset{y_{1:T}} {\mathrm{argmax}} \, p(y_{1:T} | \textbf{x}_{1:T}) $
for first order MEMMs (M=1) can be computed using the Viterbi algorithm~\cite{McCallum2000, Rabiner1989}.
In the case of higher order MEMMs we used beam search, an established approximate inference technique in natural language processing, to infer the most probable sequence~\cite{Lowerre1976}.

\subsection{Sum-Product Networks for Linear-Chain Conditional Random Fields}
\label{sec:lc-crf_cd-dhcrf}
In this section, we extend first order LC-CRFs by SPN (SPN-LC-CRFs).
\begin{figure}[ht!]
\centering
\includegraphics[scale=1.4]{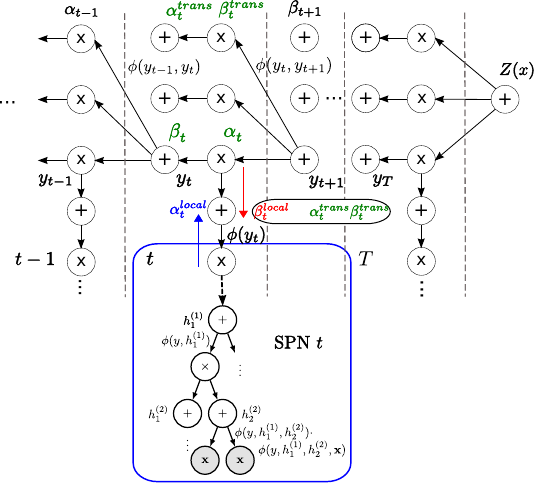}
\caption{LC-CRF extended by SPN. Illustration of the forward-backward algorithm including the computation in the SPN local factors.}
\label{fig:lc-crf_cd-dhcrf}
\end{figure}
\subsubsection{Model Definition} First order LC-CRFs model the conditional probability of sequence labels $y_{1:T}$ given a sequence of observed variables $\textbf{x}_{1:T}$ directly, i.e.
\begin{align}
  p(y_{1:T} | \textbf{x}_{1:T}) = \frac{ \prod_t \phi(y_t, \textbf{x}_{1:T}) \phi(y_{t-1}, y_t) }{Z(\textbf{x}_{1:T})},
\label{eq:pcrf}
\end{align} and
\begin{align}
  Z(\textbf{x}_{1:T}) = \sum_{y_{1:T}} \prod_t \phi(y_t, \textbf{x}_{1:T}) \phi(y_{t-1}, y_t)
\label{eq:Zcrf}
\end{align}
is the partition function.
Computation of the most probable sequence $\hat{y}_{1:T}$ can be performed using the Viterbi algorithm~\cite{Sutton2008}.

\subsubsection{Forward-backward Algorithm on the Chain} We now extend the LC-CRF by replacing these local factors $ \phi(y_t, \textbf{x}_{1:T}) = \alpha_t^{local}(y_t, \textbf{x}_{1:T})= \phi(y) \alpha^{spn}(y_t,\textbf{x}_{1:T})$ in Eq.~\eqref{eq:pcrf} and~\eqref{eq:Zcrf} by SPNs. The messages $\alpha^{spn}(y_t,\textbf{x}_{1:T})=Q(y,\textbf{x})$ are computed efficiently according to the SPN architecture using~\eqref{eq:Qsp}.
Accordingly, we adapt the forward messages
\begin{align}
\alpha_t^{trans}(y_t) = \sum_{y_{t-1}} \phi(y_{t-1},y_t) \alpha_{t-1}(y_{t-1}), \\
\alpha_t(y_t) =  \alpha_t^{local}(y_t,\textbf{x}_{1:T}) \alpha_t^{trans}(y_t)
\end{align}
 and backward messages
\begin{align}
\beta_t^{trans}(y_t) = \sum_{y_{t+1}} \phi(y_{t},y_{t+1}) \beta_{t+1}(y_{t+1}), \\
\beta_t(y_t) =  \alpha_t^{local}(y_t,\textbf{x}_{1:T}) \beta_t^{trans}(y_t),
\end{align}
where $\alpha_t^{trans}(y_t)$ and $\beta_t^{trans}(y_t)$ denote the messages passed along the linear chain without the local message $\alpha_t^{local}(y_t,\textbf{x}_{1:T})$. Further, $\alpha_t(y_t)$ and $\beta_t(y_t)$ denote the messages 
passed along the linear chain including the local message at sequence index $t$. Figure \ref{fig:lc-crf_cd-dhcrf}
shows a sum-product network representation of the forward-backward algorithm in the linear chain and how it can be extended to
deep local factors, i.e.\ SPNs.
The partial derivatives that need to be passed to the SPNs are
$\beta_t^{local}(y_t) =  \alpha_t^{trans}(y_t) \beta_t^{trans}(y_t).$
This allows for joint exact \emph{and} efficient inference and training of the LC-CRF and the SPNs in a single framework.

\section{Experiments}
\label{sec:results}

In the previous section, we derive the SPN-MEMMs and SPN-LC-CRFs only for local factors,~i.e. input-dependent factors mapping to one output label. This can be extended to higher-order (HO) input-dependent factors which can map $m$ observation vectors to $n$ consecutive labels. Due to space reasons we omit a detailed derivation. We added the term HO to the model name in such cases,~e.g. SPN-HO-LC-CRF.

\subsection{Data sets}

We evaluated the performance of the proposed models on the following two data sets:
 
\subsubsection{OCR Data Set}
The OCR data set~\cite{Taskar2004} represents an optical character recognition task.
The data set consists of 6877 handwritten words, each represented as a  sequence of handwritten characters.
These characters are provided as binary images of size $16 \times 8$ pixels and the raw pixel values serve as input features.
The task is to assign one out of 26 possible labels, i.e.\ the represented character, to each of these images. In total, 55 unique words with an average length of 8 characters are provided.
Performance is measured by the ratio of wrongly assigned labels to the total number of labels.
Furthermore, 10-fold cross-validation is used. 
The average character error rate (CER) in [\%] over all ten folds is reported.

\subsubsection{TIMIT Data Set} 
The TIMIT data set~\cite{Zue1990} contains recordings of $5.4$ hours of English speech
from 8 major dialect regions of the United States. The recordings were manually segmented at phone level. We use this segmentation for phone 
classification. Note that phone classification should not 
be confused with phone recognition~\cite{Hinton2012} where no segmentation is provided.
We collapsed the original 61 phones into 39 phones. All frames of Mel frequency cepstral coefficients (MFCCs), delta and double-delta coefficients of a phonetic segment are mapped into one feature vector.\\
The features are derived similarly as proposed in Halberstadt and Glass
(1997). First, 12
MFCC + log-energy feature (13 MFCC’s), their derivatives (13
derivatives) and their second derivatives (13
second derivatives) are calculated for
every 10ms of the utterance with a window size of 25ms. A phonetic
segment, which can be variable
length, is split at a 3:4:3 ratio into 3 parts. The fixed-length feature
vector is composed of: 1) three
averages of the 13 MFCC’s calculated from the 3 portions (39 features);
2) the 13 derivatives and the 13 second derivatives of the
beginning of the first and the end of the third segment part (26 + 26 = 52)
features); and 3) the log duration of
the segment (1 feature). Hence, each phonetic segment is represented by
92 features.\\
The task is, given an utterance and a corresponding segmentation, to infer the phoneme within every segment. The data set consists of a training set, a development set (dev) and a test set (test), containing 140.173, 50.735 and 7.211 phonetic segments, respectively.
Furthermore, the development set is used for parameter tuning. The performance measure is the phone error rate (PER) in [\%].

\subsection{Labeling Experiments}
In all experiments and for all data sets, input features were normalized to zero mean and unit standard deviation. Optimization of our models was in all cases performed using stochastic gradient ascent using a batch-size of one sample. An $\ell_2$-norm regularizer on the model weights
was used. The development set is used for hyper-parameter tuning: the learning rate $\eta \in \{ 10^{-2}, 10^{-3}, 10^{-4}\}$ and the regularization parameter $\rho \in \{ 10^{-2}, 10^{-3}, 10^{-4}\}$ for the $\ell_2$-norm regularizer are selected based on the performance on the development set. 

\subsubsection{OCR Experiments}

\begin{table}[h!t]
\small
  \caption{\emph{OCR Task:} SPN-MEMMs vs.\ SPN-LC-CRFs for various model structures ($L$, $I$, $H$). Character error rate in [\%].}
  \label{tab:ocrMEMMvesusCRF}
\vspace{-0.5em}
  \centering
  \begin{tabular}{llcccc}
   \toprule
   & & \multicolumn{2}{c}{$I=2$} &\multicolumn{2}{c}{$I=3$} \\ 
\cmidrule(lr){3-4} \cmidrule(lr){5-6}
   $L$        & & $H=2$   & $H=3$     & $H=2$   & $H=3$        \\ \midrule
   \multirow{2}{*}{1} & SPN-MEMM & 13.87 & 13.13 &  12.26 & 11.81   \\
		      & SPN-LC-CRF & 8.35 & 7.81 & 7.32 & 6.94   \\
\midrule
   \multirow{2}{*}{2} & SPN-MEMM & 10.66 & 10.40 & 9.35 & 9.37    \\
		      & SPN-LC-CRF & 6.28 & 6.53 & \textbf{5.75} & 5.76   \\
\midrule
   \multirow{2}{*}{3} & MEMM & 9.37 & \na{8.74} & \na{9.31} & n.a. \\
		      & SPN-LC-CRF & 5.77 & \na{6.76} & \na{5.87} & n.a. \\
    \bottomrule
  \end{tabular}
\vspace{-0.5em}
\end{table}

First, we compared first-order SPN-MEMMs and SPN-LC-CRFs. The experiments ran for $100$ epochs. For inference we used the Viterbi algorithm which is exact and efficient for first-order models. In Table \ref{tab:ocrMEMMvesusCRF} we explored the performance of the SPN extension for various structures,
i.e.\ different number of layers $L$, different numbers of products $I$ and different numbers of hidden states $H$.
The performance increased with increasing model size.
For similar model configurations, the SPN-LC-CRFs significantly outperformed the SPN-MEMMs.

\begin{table}[h!tp]
\small
  \caption{\emph{OCR Task:} M\textsuperscript{th} order SPN-MEMMs for different model sizes ($L=1$). Beam search with a width of $20$ has been used. Character error rate in [\%].}
  \label{tab:ocrHigherOrderMEMM}
\vspace{-0.5em}
  \centering
  \begin{tabular}{lcccc}
  \toprule
   &\multicolumn{2}{c}{$I=2$} & \multicolumn{2}{c}{$I=3$} \\ 
      \cmidrule(lr){2-3} \cmidrule(lr){4-5}
      $M-1$ & $H=2$   & $H=3$   & $H=2$   & $H=3$     \\ \midrule
 \multirow{1}{*}{1} &  14.32 & 13.81 &  12.92 & 12.39   \\
 \multirow{1}{*}{2} &  7.73 & 7.66 &  7.81 & 6.98   \\
 \multirow{1}{*}{3} &  5.64 & 5.46 &  5.27 & 5.17   \\
 \multirow{1}{*}{4} &  4.49 & 4.36 &  4.36 & 4.17   \\
 \multirow{1}{*}{5} &  3.39 & 3.99 &  3.90 & 3.80   \\
 \multirow{1}{*}{6} &  3.78 & 3.73 &  3.75 & 3.57  \\
 \multirow{1}{*}{7} &  3.73 & 3.58 &  3.60 & \na{3.36}   \\
 \multirow{1}{*}{8} &  3.69 & 3.55 &  3.61 & 3.37   \\
 \multirow{1}{*}{9} &  3.68 & 3.55 &  3.61 & \textbf{3.34}  \\
\bottomrule
  \end{tabular}
\vspace{-0.5em}
\end{table}


Second, we considered higher-order SPN-MEMMs to investigate the influence of longer label history on the performance and present these results in Table~\ref{tab:ocrHigherOrderMEMM} for different model sizes using one hidden layer.
The longer the history, i.e.\ the larger the number of previous labels $M-1$, and the larger the model, the better is the achieved performance.

\begin{table}[h!tbp]
\small
  \caption{\emph{OCR Task:} Summary. Character error rate in~[\%].}
  \label{tab:ocrSummary}
  \vspace{-0.5em}
  \centering
\begin{tabular}{rccccc}
  \toprule
  Model & &  & &  & CER [\%] \\
  \midrule
  \multicolumn{5}{l}{LC-CRF (1st order)~\cite{Maaten2011}}		& 14.2 \\
  \multicolumn{5}{l}{HU-CRF (1st order)~\cite{Maaten2011}}	& 7.73 \\
  \multicolumn{5}{l}{HU-CRF Large-margin (1st order)~\cite{Maaten2011}}	& 4.05 \\
  \multicolumn{5}{l}{HO-HU-CRF (2nd order)~\cite{Maaten2011}}		& 1.99 \\
  \multicolumn{5}{l}{Cascades LC-CRFs~\cite{Weiss2012}}		& 1.46 \\
\midrule
  \multicolumn{5}{l}{GMM-LC-CRF (1st order)} 	&   9.53 \\
  \multicolumn{5}{l}{SPN-MEMM (1st order)} 	&   9.35 \\
  \multicolumn{5}{l}{SPN-LC-CRF (1st order)} 	&   \textbf{5.75} \\

\midrule
  \multicolumn{5}{l}{SPN-HO-MEMM (higher-order)} 	&   \textbf{3.12} \\
  \multicolumn{5}{l}{SPN-HO-LC-CRF (2nd order)} 	&   \textbf{1.41} \\
  \bottomrule
\end{tabular}
\vspace{-0.5em}
\end{table}

Finally, we introduced second-order SPN-HO-LC-CRFs and summarized our best results in Table \ref{tab:ocrSummary} and compared them.
In particular, we compared our models to first-order LC-CRFs, first-order GMM-LC-CRFs (special case of our model) and first-order hidden-unit CRFs (HU-CRFs)~\cite{Maaten2011}, optimized using
stochastic gradient descent. These models achieved a labeling error of $14.2 \%$, $9.53\%$ ($L=1$, $I=1$, $H=4$) and $7.73 \%$ (hidden variables $I=250$ and states $H=2$), respectively.
All presented models are better than LC-CRFs with linear local factors,~i.e. $14.2\%$. Furthermore, for SPN-MEMM we explored one to three hidden layers for $M-1=8$ previous labels and achieved the best MEMM performance of $3.12\%$ ($L=3$, $I=2$, $H=2$). Our SPN-LC-CRFs achieved better performance $(5.70\%)$
than the first-order HU-CRFs and first-order GMM-LC-CRF.
Our best result $(1.41\%)$ has been achieved with the second-order SPN-HO-LC-CRFs ($L=3$, $I=2$, $H=4$) and outperformed the second-order
HU-CRF $(1.99\%)$~\cite{Maaten2011} and the Cascades of LC-CRFs $(1.46\%)$~\cite{Weiss2012}.

\subsubsection{TIMIT Experiments}

\begin{table*}[h!t]
\small
  \caption{\emph{TIMIT Task:} First-order SPN-MEMMs vs.\ SPN-LC-CRFs for various model structures. Phone error rate in [\%].}
  \label{tab:timitCRF}
\vspace{-0.5em}
  \centering
  \begin{tabular}{llccccccccc}
  \toprule
    & & \multicolumn{3}{c}{$I=2$} & \multicolumn{3}{c}{$I=3$} & \multicolumn{3}{c}{$I=4$} \\ \cmidrule(lr){3-5} \cmidrule(lr){6-8} \cmidrule(rl){9-11}
      SPN-MEMM      & & $H=2$   & $H=3$   & $H=4$   & $H=2$   & $H=3$   & $H=4$   & $H=2$   & $H=3$   & $H=4$   \\ \midrule
   \multirow{2}{*}{$L=1$} & dev & 23.04 & 22.30 & 21.88 & 22.35 & 21.68 & 22.52 & 22.21 & 21.44 & 21.30 \\
                           & test & 23.58 & 22.97 & 22.70 & 23.30 & 22.60 & 22.25 & 23.20 & 22.26 & 22.40 \\[0.2cm]
   \multirow{2}{*}{$L=2$} & dev & 21.58 & 21.11 & 21.06 & 21.01 & \textbf{20.55} & 20.60 & 21.29 & 20.84 & 20.62 \\
                           & test & 22.18 & 22.13 & 21.97 & 22.02 & \textbf{22.15} & 21.55 & 22.78 & 22.22 & 21.96 \\
\midrule
SPN-LC-CRF   & & && & && & && \\
   \multirow{2}{*}{$L=1$} & dev & 21.92 & 21.21 & 21.03 & 21.20 & 20.77 & 20.69 & 21.21 & 20.52 & 20.29  \\
                           & test & 22.40 & 22.00 & 21.90 & 21.96 & 21.52 & 21.03 & 21.71 & 20.92 & 20.73  \\[0.2cm]

   \multirow{2}{*}{$L=2$} & dev & 20.51 & 20.24 & 20.23 & 20.23 & 19.92 & \textbf{19.88} & 20.37 & 20.05 & 19.89 \\
                           & test & 21.08 & 20.98 & 20.74 & 21.14 & 21.17 & \textbf{20.90} & 21.75 & 21.66 & 20.93 \\ \bottomrule
  \end{tabular}
\vspace{-0.5em}
\end{table*}


\begin{table}[h!t]
\small
  \caption{\emph{TIMIT Task:} Sum-product network higher-order LC-CRF (SPN-HO-LC-CRF). Phone error rate in [\%].}
  \label{tab:timitSPNCRF}
\vspace{-0.5em}
  \centering
  \begin{tabular}{llcccc}
  \toprule
    & & \multicolumn{2}{c}{$I=2$} & \multicolumn{2}{c}{$I=3$}  \\ 
\cmidrule(lr){3-4} \cmidrule(lr){5-6}       & & $H=2$   & $H=3$      & $H=2$   & $H=3$ \\ 
\midrule
   \multirow{2}{*}{$L=1$} & dev & 19.6 & 18.0 & 18.6 &  18.1 \\
                           & test & 19.7 & 18.3 & 19.5 & 18.8 \\[0.2cm]
   \multirow{2}{*}{$L=2$} & dev & 18.1 & 17.3 & 18.7 &  18.1 \\
                           & test & 18.7 & 18.3 & 19.5 & 18.5 \\[0.2cm]
   \multirow{2}{*}{$L=3$} & dev & 18.0 & 17.3 & 19.0 &  18.0 \\
                           & test & 18.5 & \textbf{18.2} & 20.0 & 18.5 \\[0.2cm]

\bottomrule
  \end{tabular}
\vspace{-0.5em}
\end{table}

\begin{table}[h!t]
\small
  \caption{\emph{TIMIT Task:} Higher-order hidden unit CRF (HO-HU-CRF). Phone error rate in [\%].}
  \label{tab:timitHUCRF}
\vspace{-0.5em}
  \centering
  \begin{tabular}{llcccc}
  \toprule
    & & \multicolumn{2}{c}{$H=150$} & \multicolumn{2}{c}{$H=200$}  \\
\midrule
   \multirow{2}{*}{\,\,$m=n=1$} & dev & \na{19.6} &  & 19.3  &    \\
                           & test & \na{20.5} &   & 20.6  &  \\[0.2cm]
   \multirow{2}{*}{+\,$m=n=2$} & dev & \na{17.3}  &  & 17.3  &    \\
                           & test & \na{17.8}  &   & 18.0  &  \\[0.2cm]
\multirow{2}{*}{+\,$m=n=3$} & dev & \na{19.7}  &   & \na{20.0}  &    \\
                           & test & \na{20.5}  &  & \na{20.8}  &   \\[0.2cm]
\bottomrule
  \end{tabular}
\vspace{-0.5em}
\end{table}

We report the test performance corresponding to the best performance on the development set during $500$ training epochs.
Detailed results for our first-order SPN-MEMMs and SPN-LC-CRFs on the development set as well as the core-test set are provided in Table~\ref{tab:timitCRF}. We explored various structures of the SPN ($L\in\{1,2,3\}$, $I\in\{2,3,4\}$, $H\in\{2,3,4\}$). SPN-LC-CRFs outperformed SPN-MEMMs. Larger model sizes improved performance.\\
In the next set of experiments, we extended our LC-CRFs to higher-order factors. We used linear unigram, bigram and trigram features in the input-independent factors.
In addition to local factors that map $m$ consecutive observation vectors ($m$ segments) to one label, we used also higher-order input-dependent factors that map $m$ observation vectors to $n$ consecutive labels. Higher-order factors represented as MLP networks have been already used in~\cite{Ratajczak2015, Ratajczak2015a}. In this work, we represent the input-dependent higher-order factors by SPNs and for comparison by discriminative RBMs (higher-order HU-CRFs). To the best of our knowledge, we use SPNs and RBMs for the first time to model input-dependent higher-order factors in LC-CRFs. So, we tested input-dependent factors with $m=n=1$, $m=n=2$ and $m=n=3$ in addition to the input-independent bigrams and trigrams. In preliminary experiments with SPN-HO-LC-CRFs and HO-HU-CRFs we found that using input-dependent factors $m=n=3$ leads to over-fitting. We also observed that over-fitting was reduced by using sparse input-dependent and input-independent 
factors, i.e.\ we used only bigram factors and trigram factors which have been observed in the training data at least once. So in the following experiments, we used only input-dependent factors $m=n=1$ and $m=n=2$. Both input-dependent and input-independent factors used sparse bigram and trigram factors. For this configuration we replaced the summations by a maximum operator. This might improve the performance similar as in~\cite{Gens2012}. We observed that using summations in SPN-HO-LC-CRF (L=1, I=2, H=2) gave slightly better performance than the maximum operator on the development and test set, i.e.\ we observed a performance of $19.6\%$ and $19.7\%$ compared to a performance of $19.9\%$ and $19.9\%$, respectively. In Table~\ref{tab:timitSPNCRF}, we summarized the results for SPN-HO-LC-CRFs. We experimented with one to three hidden layers and different numbers of products $I$, as well as different numbers of hidden states $H$ ($L\in\{1,2,3\}$, $I\in\{2,3\}$, $H\in\{2,3\}$). We achieved our best 
performance of $18.2\%$ with the SPN-HO-LC-CRFs (L=3, I=2, H=3).\\
In Table~\ref{tab:timitHUCRF}, we present results for HO-HU-CRFs using different numbers of hidden units $I\in\{150,200\}$, binary states $H=2$ and different orders $m=n$ of the input-dependent factors. The plus sign in Table~\ref{tab:timitHUCRF} indicates that the input-dependent factors of lower-order are also included. The best performance of $17.8\%$ for HO-HU-CRFs is slightly better than that of SPN-HO-LC-CRFs, however, SPN-HO-LC-CRFs are still competitive.\\
%
\begin{table}[h!tb]
\small
  \caption{\emph{TIMIT Task:} Summary of labeling results. Performance measure: Phone error rate (PER) in [\%].} 
  \label{tab:timit}
  \vspace{-0.5em}
  \centering
\begin{tabular}{rcccc}
  \toprule
  Model &  &  &  & PER [\%] \\
  \midrule
  \multicolumn{4}{l}{GMMs ML~\cite{Sha2006}} & {25.9} \\
  \multicolumn{4}{l}{HCRFs~\cite{Sung2007}} & {21.5} \\
  \multicolumn{4}{l}{Large-Margin GMM~\cite{Sha2006}} & {21.1} \\
  \multicolumn{4}{l}{Heterogeneous Measurements~\cite{Halberstadt1997}} & {21.0} \\
  \multicolumn{4}{l}{CNF; 1 seg.} & {20.67} \\
\midrule
  \multicolumn{4}{l}{GMM-LC-CRF (1st order); 1 seg.} & {22.72}\\
  \multicolumn{4}{l}{GMM-LC-CRF (1st order) diag; 1 seg.} & {24.21}\\
  \multicolumn{4}{l}{GMM-LC-CRF (1st order); 3 seg.} & {22.10}\\
\midrule
  \multicolumn{4}{l}{SPN-MEMM (8th order); 1 seg.} & {22.15}\\
  \multicolumn{4}{l}{SPN-LC-CRF (1st order); 1 seg.} & {20.54}\\
  \multicolumn{4}{l}{SPN-LC-CRF (1st order); 3 seg.} & {19.95}\\
\midrule
  \multicolumn{4}{l}{SPN-HO-LC-CRF (2nd order)} & {\textbf{18.2}}\\
  \multicolumn{4}{l}{HO-HU-CRF (2nd order)} & {\na{\textbf{17.8}}}\\
\multicolumn{4}{l}{NHO-LC-CRF (2nd order) \cite{Ratajczak2015a}} & {\textbf{17.7}}\\
  \bottomrule
\end{tabular}
\vspace{-0.5em}
\end{table}
\noindent Finally, we summarize our results in Table~\ref{tab:timit} and compare it to other state-of-the-art 
methods, namely hidden conditional random fields (HCRFs)~\cite{Sung2007},
large-margin GMMs~\cite{Sha2006}, heterogeneous measurements~\cite{Halberstadt1997} and CNFs~\cite{Peng2009}.
We also considered conditional neural fields (CNFs) which combine LC-CRFs with multi-layer neural networks. Using the software of~\cite{Peng2009} we tested CNFs with $50$, $100$ and $200$ hidden units as well as one and three input segments. We achieved the best result with 100 hidden units and one segment as input (1 seg.).
Large-margin GMMs outperformed generative GMMs and LC-CRFs augmented by GMMs.
However, our best first-order SPN-LC-CRFs using 3 segments as input already outperformed the other state-of-the-art methods.
Furthermore, our best SPN-HO-LC-CRFs and HO-HU-CRFs achieved even better performance of $18.2\%$ and $17.8\%$, respectively.
These results compare well to the performance of $17.7\%$ for NHO-LC-CRFs~\cite{Ratajczak2015} using MLP networks as higher-order factors (2\textsuperscript{nd} order) up to $m=n=3$ input-dependent factors.


%


\section{Discussion and Future Work}
\label{sec:conclusion}

We considered sum-product networks (SPNs) enabling both exact \emph{and} efficient inference. Furthermore, we extended linear-chain CRFs and maximum 
entropy Markov models (MEMMs) by replacing local factors with SPNs. 
Finally, we empirically evaluated our models for sequence labeling. Results for phone classification and optical character recognition are provided and are competitive in all cases.
\\
In future work, we plan to extend our model to phone recognition by using segmental LC-CRFs \cite{Zweig2009}. Furthermore, we aim at exploiting the possibility to easily calculate the marginals of the hidden variables in our model for applying posterior constraints to the hidden variables~\cite{Ganchev2010}. 

\ifCLASSOPTIONcaptionsoff
  \newpage
\fi



\bibliographystyle{IEEEtran}
\bibliography{IEEEabrv,icml2014}




\end{document}